# SwaQuAD-24: QA Benchmark Dataset in Swahili

Kondoro Alfred Malengo[1]

[1]Department of Data Science, Hanyang University, Seoul, South Korea
alfr3do@hanyang.ac.kr

## Abstract

*This paper proposes the creation of a Swahili Question Answering (QA) benchmark dataset, aimed at addressing the underrepresentation of Swahili in natural language processing (NLP). Drawing from established benchmarks like SQuAD, GLUE, KenSwQuAD, and KLUE, the dataset will focus on providing high-quality, annotated question-answer pairs that capture the linguistic diversity and complexity of Swahili. The dataset is designed to support a variety of applications, including machine translation, information retrieval, and social services like healthcare chatbots. Ethical considerations, such as data privacy, bias mitigation, and inclusivity, are central to the dataset's development. Additionally, the paper outlines future expansion plans to include domain-specific content, multimodal integration, and broader crowdsourcing efforts. The Swahili QA dataset aims to foster technological innovation in East Africa and provide an essential resource for NLP research and applications in low-resource languages.*

## 1. Introduction

The East Africa region boasts a rich Swahili linguistic heritage, with the language being spoken by millions across the region [1]. Tanzania promoted Swahili to national language status in favour of other ethnic languages as part of efforts to foster national unity. Political and ideological influence on both formal and informal development of the language in Tanzania is a major factor to consider [2]. Due to its versatility, Swahili has been used as a medium of instruction and communication in religion which has led to it adapting religious words from Islam or Christianity. Swahili has a huge potential to further influence integration and development especially in East Africa[3].

Natural language processing (NLP) has made significant strides in recent years with the development of benchmarks like GLUE, SQuAD, and KLUE driving progress in language understanding tasks. These benchmarks provide a foundation for training and evaluating models on tasks such as question answering (QA), sentiment analysis, and natural language inference. However, the majority of these benchmarks and datasets focus on high-resource languages like English, leaving lower-resource languages like Swahili underrepresented [4] [1]. Despite being spoken by over 100 million people across East Africa as a lingua franca,there is a lack of substantial resources in the NLP field particularly in the area of QA datasets.

GLUE and SQuAD are widely used datasets for the English language. GLUE is a collection of diverse natural language understanding tasks, including sentiment analysis, textual entailment, and linguistic acceptability, that serve as a comprehensive evaluation of a model's language understanding capabilities [5]. SQuAD, on the other hand, is a popular question-answering dataset that challenges models to answer questions based on given passages of text [6].

Similarly, KLUE was created for the Korean language, providing a similar suite of language understanding tasks to assess the performance of Korean language models[7]. These high-quality, multi-task datasets have become the de facto standards for benchmarking and driving progress in natural language processing research. The availability of such comprehensive, multi-task datasets has been instrumental in advancing the field of natural language processing, as they provide a standardised and rigorous framework for evaluating the performance of language models on a diverse range of tasks. By establishing these benchmarks, the research community can more effectively compare and improve language models, leading to significant advancements in the understanding and processing of natural language.

However, the availability of similar comprehensive datasets for low-resource languages, such as Swahili, remains limited. In this paper, I propose the development of SwaLUE a Swahili language understanding evaluation dataset that aims to provide a robust benchmark for assessing the performance of Swahili natural language processing models. This dataset will serve as a critical resource to address this gap and support the advancement of NLP tools for Swahili-speaking communities.

## 2. Problem Statement

While natural language processing (NLP) has advanced significantly for high-resource languages like English, Spanish, and French, low-resource languages such as Swahili remain underrepresented in NLP research. The lack of large-scale datasets for evaluating Swahili NLP models is a significant barrier to the development of language technologies for this important language. Some of the selected notable existing datasets including the Kencorpus project have collected text and speech data for Swahili, Dholuo, and Luhya language[8], Sawa Corpora which is a two-million-word parallel corpus of English-Swahili developed to address the difficulties of finding appropriate and easily accessible data for the English-Swahili language pair[9] and KenSwQuAD, a question answering dataset based on Swahili language stories[10]. However, due to the quick improvement of Large Language Models (LLMs) and task-specific datasets there is a need for a more comprehensive Swahili benchmark to drive progress[11].

The need for a Swahili QA benchmark dataset is critical. Such a dataset would enable the development of advanced question-answering systems that could significantly benefit Swahili-speaking communities by improving access to information, enhancing educational resources, and supporting a wide range of applications[10][11]. Existing Swahili QA datasets like KenSwQuAD have demonstrated the potential of such resources, but a more comprehensive, multi-task benchmark is needed to fully unlock the potential of Swahili natural language processing and in turn bridge the gap between low- and high-resource languages in the NLP space. Additionally, the creation of a diverse QA dataset covering multiple domains, question types, and dialectal variations of Swahili would provide the groundwork for more accurate and reliable language models, benefiting millions of Swahili speakers across Africa.

## 3. Proposed Solution: Swahili QA Benchmark Dataset

### 3.1. Dataset Format

This benchmark will consist of multiple types of question-answer pairs to simulate varied QA tasks. These formats will include:

- Reading comprehension: Questions based on provided Swahili passages that require answers derived from the text.
- Multiple choice: Questions with a set of answer options, where the correct answer must be selected. This will be designed to evaluate understanding of the text by providing several potential answers to each question.
- Factoid QA: Simple questions that can be answered with a word or short phrase, such as named entities, dates, or numerical information.

- Unanswerable Questions: As seen in datasets like SQuAD 2.0, some questions will have no corresponding answers in the provided text, testing the model's ability to detect insufficient information.
- Freeform QA: Open-ended questions that require a more detailed, sentence-level response.

Additionally, the dataset will provide metadata for each question-answer pair, including the source passage, difficulty level, question type, and other relevant information to facilitate in-depth analysis and targeted model evaluations.

### 3.2. Data Sources

The creation of the Swahili QA benchmark dataset will involve sourcing data from a variety of text types to ensure diversity in both language style and domain coverage. Hence the plan is to curate content from a diverse set of high-quality Swahili-language sources, potentially including:

- Primary Data: This will be collected directly from Swahili Speakers and communities. In addition, as employed in [10], story writing competitions can be a key avenue for gathering diverse, informal text from a broad range of authors.
- Secondary Data: Existing Swahili language datasets such as Kencorpus, KenSwQuAD, and Sawa Corpora will be leveraged to supplement the primary data collection efforts. published books, news articles, government publications, and media houses. These sources will provide more formal, structured language and will cover topics ranging from current affairs to academic subjects [10] [11].
- Crowd-sourcing: Furthermore, some of the content and question-answer pairs may be generated through carefully designed crowd-sourcing initiatives, allowing for broader community participation and input. The resulting dataset will be thoroughly reviewed and validated by Swahili language experts to ensure the highest possible quality and accuracy before release.
- Digital Content: To capture modern Swahili usage, web content such as blog posts, discussion forums, and social media platforms will also be mined. This will introduce informal language variations, colloquialisms, and mixed language use, reflecting Swahili in digital spaces.
- Translated Data: Select English datasets like SQuAD, TyDiQA, or other relevant QA benchmarks can also be considered for translation to Swahili, ensuring the inclusion of a broader range of question

### 3.3. Benchmarking and Evaluation

In recent years, question answering (QA) has become one fo the most critical benchmarks for assessing language understanding and generation capabilities of AI systems across a variety of languages. The development of high-resource language benchmarks such as SQuAD (Stanford Question Answering Dataset) and TyDiQA(A Multilingual Question Answering Benchmark)[6][12] has paved the way for significant advance in NLP, providing structured datasets for evaluating model performance on diverse question types. These benchmarks have contributed to state-of-the-art (SOTA) language models, such as BERT, RoBERTa, and their multilingual variants (e.g., mBERT and XLM-R), through rigorous evaluations based on metrics like F1 score and Exact Match (EM).

However, while these benchmarks have been instrumental in advancing QA models, they have been disproportionately focused on languages with abundant resources, such as English, Chinese, and French. TyDiQA expanded the scope by covering low-resource languages, yet

Swahili's inclusion in NLP tasks remains sparse and insufficient for training models robust enough for practical applications in East Africa. Consequently, there is an urgent need for a comprehensive Swahili QA benchmark that can rigorously assess models for their ability to understand and generate accurate answers in Swahili. This benchmark will adopt proven evaluation metrics, such as F1 and EM, but adapt them to the specific linguistic properties of Swahili.

### 3.3.1. Evaluation Metrics

- Previous Use: In datasets such as [12] [11] [13], Exact Match(EM) measures the percentage of predictions where the system's response exactly matches the correct answer from the dataset text. This metric is essential for fact-based QA where answers require high precision.
- Importance for Swahili: For the Swahili dataset, EM will capture the ability of models to generate precise answers despite the language's rich morphological structure. Swahili, with its extensive use of noun classes and affixes, presents challenges where models must output the exact form, including verb tense and noun plurality. For example, in a task requiring models to identify the plural form of "mtoto" (child), the correct answer "watoto" (children) must exactly match the gold-standard label. The use of EM will help evaluate model performance in these syntactically sensitive areas.
- F1 Score

    - Previous Use: Widely used alongside EM, the F1 score allows for partial credit by measuring the overlap between the predicted and correct answers, balancing precision and recall. This was critical in benchmarks like SQuAD, where short phrase answers needed to be identified with longer context passages.
    - Importance of Swahili: Given Swahili's agglutinative nature, where words can vary due to affixation or morphology without a change in meaning, F1 will be especially relevant. For instance, if a model predicts "alikuwa na kitabu" (he had a book) instead of "alishika kitabu" (he held the book), the overlap in meaning would still be high, and F1 ensures the model is not overly penalised for such close predictions. F1 will be particularly effective for free-form answers or reading comprehension tasks, where there may be several acceptable ways to phrase the correct response.

- BLEU and Perplexity

    - Previous Use: In language generation tasks, BLEU and perplexity are common metrics for evaluating the quality and fluency of model outputs. BLEU is a metric used to evaluate the quality and fluency of model-generated text by measuring the overlap between the generated sentences and one or more reference sentences. It assesses the precision of n-grams, or sequences of words, within the model output, with a higher BLEU score indicating better alignment with the reference text and, thus, higher-quality generation performance. On the other hand, Perplexity is a metric that evaluates how well a model predicts each word in a sequence based on the preceding context, without directly comparing the generated text to a reference. This provides an assessment of the model's ability to generate fluent and coherent language, complementing the reference-based evaluation offered by BLEU.

- Importance in Swahili: As part of the proposed Swahili QA dataset, models might need to generate fluent, contextually appropriate responses, especially in open-ended or narrative QA tasks. BLEU will be crucial in evaluating how well a model can generate answers that syntactically and semantically resemble human-provided answers, capturing nuances like affixation and word order. Perplexity will help assess the fluency and grammatical correctness of responses, ensuring that generated Swahili text adheres to natural language use patterns.
- Additional Metrics for Low-Resource Contexts:
  - Matthews Correlation Coefficient (MCC): Introduced in the GLUE benchmark[5], MCC evaluates binary classification tasks, like determining if a sentence is grammatically acceptable. MCC could be adapted for QA to evaluate binary classification tasks such as fact verification or entailment in Swahili, ensuring balanced performance even in cases of class imbalance.
  - Macro F1 and AUC-PR: These metrics, as used in KLUE[7], are essential for evaluating tasks like relation extraction or named entity recognition in low-resource languages. In the Swahili dataset, similar metrics could be employed to assess tasks like multi-class classification where more nuanced relational understanding is required between entities.

### 3.3.2. Baseline Models

- Multilingual Baseline Models (mBERT, XLM-R); The rich success of multilingual models like mBERT and XLM-R in zero-shot and few-shot learning on low-resource languages [14] suggests that these foundational models can serve as a strong starting point for Swahili QA.
  - Previous Use: Multilingual models such as mBERT and XLM-R have been foundational baselines in datasets like TyDiQA and KenSwQuAD, enabling cross-lingual transfer learning. Despite their success in handling multiple languages, their performance on low-resource languages like Swahili has shown limitations, particularly due to language-specific nuances they were not fine-tuned on.
  - Importance for Swahili: In the Swahili QA benchmark, mBERT and XLM-R will serve as initial baselines. Their performance will highlight areas where multilingual models excel and where they struggle with Swahili's rich morphology. A thorough evaluation of these models will guide the future development of Swahili-specific fine-tuned models, contributing to improved QA systems for the language.
- Swahili-Specific Models: To fully leverage the unique characteristics of Swahili, specialised models fine-tuned on Swahili data will be essential.
  - Challenges: Current pre-trained models have limited exposure to Swahili data, and this gap in representation could lead to performance degradation when compared to high-resource languages. This presents an opportunity for fine-tuning on the Swahili QA dataset to improve baseline performance.
  - Future Prospects: Despite high accuracy of SwahBERT compared to mBERT, the performance is still suboptimal on downstream tasks[15]. With the introduction of the proposed Swahili QA benchmark, further fine-tuning and customization of Swahili-specific models can lead to substantial improvements, pushing the state-of-the-art for Swahili QA systems.

### 3.4. Ethical Considerations

Inspired by the ethical frameworks employed in previous QA benchmarks like KLUE, the development of the Swahili QA dataset will prioritise ethical considerations to ensure that the dataset is fair, inclusive, and representative of the Swahili-speaking population. The following sections outline how the dataset will address key ethical issues:

#### 3.4.1. Data Privacy

One of the primary concerns in creating a QA dataset, especially one targeting low-resource languages like Swahili, is maintaining the privacy of the individuals whose data might be included. Following the example set by KLUE, the Swahili QA dataset will adhere to strict privacy standards to ensure that no personally identifiable information (PII) is included in the dataset. This is particularly important when collecting data from sources such as blogs, social media, or community-generated content, which might inadvertently contain sensitive details about individuals from Swahili-speaking regions. Measures will include:

- Anonymizing data: Personal information such as names, locations, or identifiable descriptions will be anonymized or excluded during preprocessing.
- Compliance with regulations: The dataset will comply with local and international data protection laws, such as the General Data Protection Regulation (GDPR) or local regulations in East African countries, ensuring that individuals' data privacy rights are respected.
- Informed consent: For any data collected directly from individuals, the dataset will obtain explicit consent for the use of their information in the QA benchmark.

#### 3.4.2. Bias and Fairness

The Swahili language is spoken across a diverse range of geographical regions, socioeconomic backgrounds, and cultural contexts. To ensure that the QA dataset is representative of this diversity, the data collection process will prioritise the inclusion of a wide range of perspectives and experiences. Swahili is spoken in several dialects across Tanzania, Kenya, Uganda, the Democratic Republic of the Congo, and beyond, with linguistic variations based on region and social factors. To prevent bias:

- Dialectal Diversity: The dataset will include text from a variety of Swahili dialects, ensuring representation from different regions, such as coastal Swahili, Kenyan Swahili, and Congolese Swahili. This avoids over-representing any single variant of the language and ensures a broader understanding of Swahili.
- Demographic Representation: The dataset will strive to include content generated by individuals from diverse backgrounds, considering factors such as gender, age, socioeconomic status, and educational level.
- Inclusive Language: The dataset will be carefully curated to avoid biases, stereotypes, or offensive language that might marginalise or misrepresent certain groups within the Swahili-speaking community.
- Gender Inclusivity: Special care will be taken to balance gender representation in the texts and questions generated, ensuring that the content does not favour gender-specific roles or assumptions. Drawing on methods from KenSwQuAD, which emphasised diversity in their annotation process, the Swahili QA dataset will ensure that questions

- and answers reflect both male and female perspectives, especially in culturally sensitive topics.
- Eliminating Ethnic and Cultural Bias: Like KLUE's ethical considerations, which explicitly avoid social biases, the Swahili QA dataset will remove any examples that could reinforce stereotypes or propagate ethnic, cultural, or social biases. Care will be taken to scrutinise the data for implicit biases during the annotation process, with annotators trained to identify and flag potentially biased content.

### 3.4.3. Annotation Quality and Accuracy

High-quality annotations are crucial for the success of any QA dataset, as they directly affect the validity of the evaluation metrics like F1 and BLEU. Inspired by the rigorous annotation protocols followed by KenSwQuAD and KLUE, the Swahili QA dataset will adopt a structured annotation process that includes:

- Crowdsourcing and Expert Review: Similar to the practices used in SQuAD and KenSwQuAD, the Swahili QA dataset will be developed using a combination of crowd-sourced annotations and expert reviews. Native Swahili speakers from diverse regions will be employed as annotators to ensure linguistic and cultural authenticity.
- Training Annotators: Annotators will be thoroughly trained to avoid ambiguity in formulating question-answer pairs. This is essential in Swahili, where minor grammatical variations can significantly alter meaning. Inter-annotator agreement will be closely monitored to maintain consistency across the dataset, and regular calibration sessions will be conducted to refine guidelines and resolve any disagreements.
- Rigorous Quality Control: A sample of the dataset (similar to the 12.5% quality assurance check in KenSwQuAD) will undergo thorough verification by experts to ensure that questions are relevant, unambiguous, and that answers are correctly matched. This will ensure that the data used for training models is of the highest possible quality.
- Sourcing Permissions: Following the model of datasets like SQuAD, which use publicly available data (e.g., Wikipedia articles), the Swahili QA dataset will prioritise sources that are open access or for which appropriate permissions have been obtained. Where data is collected from individuals or institutions, informed consent will be secured, ensuring ethical sourcing practices.
- Inclusivity in Topics: The dataset will include questions and answers on a wide range of topics, ensuring that the content is not skewed toward any particular domain, such as urban-focused or elite-oriented topics. This aligns with the inclusivity demonstrated in KLUE's diverse topic selection, which encompassed news, dialogue systems, and reviews. The Swahili QA dataset will similarly include varied sources such as news articles, folk stories, educational materials, and informal social media texts, ensuring that it represents the full spectrum of Swahili usage in daily life.

## 4. IMPACT AND FUTURE WORK

### 4.1. Potential Applications

The creation of the Swahili QA benchmark dataset promises significant advancements in various domains of research and industry. Question Answering (QA) has become a pivotal component in NLP, powering applications such as search engines, chatbots, and virtual assistants. By providing a high-quality QA dataset for Swahili, this project will have a broad impact across several areas:

- Machine Translation and Cross-Lingual Understanding: The Swahili QA dataset can be integrated with multilingual NLP models, enhancing machine translation tools like Google Translate, making them more accurate in translating Swahili. As seen with cross-lingual benchmarks like TyDiQA and XGLUE, training models on QA datasets improves their ability to understand context, handle idiomatic expressions, and capture the nuances of language-specific constructions. For instance, TyDiQA demonstrated that multilingual models improve when fine-tuned on QA datasets in low-resource languages, which could similarly benefit Swahili by enhancing translation and bilingual text retrieval tasks [12][16].
- Information Retrieval and Search: Just as datasets like SQuAD revolutionised English-based information retrieval systems by enhancing models' comprehension of text, a Swahili QA dataset can be used to build Swahili-based search engines that provide more accurate and contextually relevant results. With Swahili being one of the most widely spoken languages in Africa, this dataset can empower industries particularly in East Africa to offer more localised search capabilities, benefiting education, media, and e-commerce sectors.
- Education and Language Learning: As demonstrated by GLUE, benchmark datasets enable developers to train models for use in educational tools such as language learning apps, which are particularly useful for low-resource languages. The Swahili QA dataset will be critical for developing intelligent tutoring systems and language-learning apps that focus on Swahili comprehension. In addition, educational platforms can use these datasets to offer interactive reading comprehension tasks, supporting both native Swahili speakers and learners.
- Social Services and Chatbots: Inspired by the rise of intelligent question-answering systems in healthcare and customer service, this dataset can facilitate the development of Swahili-speaking virtual assistants and chatbots. These applications can help bridge language barriers in healthcare, legal services, and government offices by providing conversational AI systems that understand and respond to user inquiries in Swahili. With a Swahili-speaking population exceeding 100 million people across Africa, the potential social impact is enormous.

### 4.2. Future Expansion

While the initial version of the Swahili QA benchmark dataset will cover essential domains such as literature, news, and educational texts, future expansions are envisioned to make the dataset more comprehensive and robust:

- Domain Expansion: Future versions of the dataset will expand to cover additional specialised domains, including healthcare, legal texts, and government documents. This aligns with the approach of KenSwQuAD, which collected text from diverse genres, ensuring coverage across domains of high social importance. By adding these specialised areas, the Swahili QA dataset can support more domain-specific applications, such as legal chatbots and medical question-answering systems.
- Multimodal QA Integration: Another avenue for future growth involves integrating multimodal data, such as pairing questions and answers with images or videos, as seen in datasets like Visual Question Answering (VQA)[17]. This could enhance the dataset's ability to support applications in areas like visual storytelling or e-learning, where users might ask questions related to visual content (e.g., diagrams in textbooks or images in online articles).
- Crowdsourcing and Collaborative Development: Building on the success of large crowdsourced efforts such as SQuAD and TyDiQA, the expansion of the Swahili QA dataset will involve a broader base of annotators from Swahili-speaking communities across East Africa. Crowdsourcing will help to increase the dataset's size and diversity, ensuring that it captures the full range of dialectal and cultural variation in Swahili. Additionally, collaboration with universities and linguistic institutions across Africa will ensure academic rigour and authenticity in the expansion process.
- Collaborations and Funding: As with other low-resource language projects, international funding bodies like the Lacuna Fund (which funded KenSwQuAD) or Digital Public Goods Alliance could be targeted for collaboration to support future expansions. By securing partnerships with institutions interested in language preservation and technology, the Swahili QA dataset can continue to grow, both in terms of scope and quality. Further, collaboration with organisations focused on AI ethics will ensure that the dataset remains ethically sound and fair as it expands.

## 5. CONCLUSIONS

The Swahili QA benchmark dataset is a vital step toward addressing the underrepresentation of low-resource languages in natural language processing (NLP). Drawing on the success of benchmarks like SQuAD, GLUE, KenSwQuAD, and KLUE, this dataset will provide the foundation for developing accurate, culturally relevant AI systems tailored to the linguistic complexities of Swahili. Its diverse and ethically curated content will empower models to improve cross-lingual understanding, machine translation, information retrieval, and educational tools, while supporting critical social services through virtual assistants and chatbots. By incorporating strong ethical standards, including data privacy, bias mitigation, and dialectal diversity, this dataset will ensure fair representation and reliability. Future expansions to cover specialised domains, integrate multimodal content, and involve broader crowdsourcing will further solidify the dataset's impact on both research and practical applications. Ultimately, the Swahili QA dataset is positioned to foster technological innovation across East Africa and beyond, contributing to the global effort to build inclusive, high-quality NLP tools for underrepresented languages.